\documentclass[12 pt]{article} 
\usepackage[margin=2.5cm]{geometry}
\usepackage{microtype}

\usepackage{graphics} 
\usepackage{epsfig} 
\usepackage{mathptmx} 
\usepackage{times} 
\usepackage{amsmath} 
\usepackage{tikz-cd}
\usepackage{amssymb}  
\usepackage{import}
\usepackage{algorithm}
\usepackage{algpseudocode}
\usepackage{multirow}
\usepackage{multicol}
\newcommand{\dimx}[0]{p}
\newcommand{\dimu}[0]{q}
\newcommand{\dimw}[0]{r}
\newcommand{\dimy}[0]{l}

\newcommand{\trace}[0]{Tr}

\newcommand{\dL}[1]{\frac{\partial \mathcal{L}}{\partial #1}}
\newcommand{\fracd}[2]{\frac{\partial #1}{\partial #2}}
\newcommand{\R}{\mathbb{R}}

\title{\LARGE \bf
Backpropagation-Based Analytical Derivatives   of EKF  Covariance for Active Sensing
}

\usepackage{authblk}
\author[1,2]{Jonas Benhamou}
\author[1]{Silvère Bonnabel}
\author[2]{Camille Chapdelaine}

\affil[1]{\normalsize Mines Paris, PSL Research University, Centre for Robotics, 60 bd Saint-Michel, 75006 Paris, France, silvere.bonnabel@minesparis.psl.eu}
\affil[2]{\normalsize SAFRAN TECH, Groupe Safran, Rue des Jeunes Bois - Chateaufort, 78772 Magny Les Hameaux CEDEX, France, \{camille.chapdelaine, jonas.benhamou\}@safrangroup.com}

\begin{document}
\maketitle
\thispagestyle{empty}
\pagestyle{empty}

\begin{abstract}
To enhance accuracy of robot state estimation, active sensing (or perception-aware) methods seek trajectories that maximize the information gathered by the sensors. To this aim, one possibility is to seek trajectories that minimize the (estimation error) covariance matrix  output by an extended Kalman filter (EKF), w.r.t. its control inputs over a given horizon.  However, this is computationally demanding. In this article, we derive novel backpropagation analytical formulas for the derivatives of the  covariance matrices of an EKF w.r.t. all its inputs. We then leverage the obtained analytical  gradients as an enabling technology to derive perception-aware optimal motion plans.   Simulations  validate the  approach, showcasing  improvements in   execution time, notably over PyTorch's automatic differentiation. Experimental results on a real     vehicle   also  support the method. \\

Keywords:
perception-aware, extended Kalman filter, trajectory optimization, backpropagation.
\end{abstract}

\section{Introduction}
\label{sec:Introduction}

In robotics,  perception-aware (PA) approaches, \cite{bartolomei_perception-aware_2020, costante_perception-aware_2017, murali_perception-aware_2019, xing_autonomous_2023}, or  active sensing approaches, seek trajectories that maximize information gathered from sensors so as  to perform robotic tasks safely.  Notably, in the context of ground vehicles, when localization is based on ranging or bearing measurements relative to beacons, the efficiency of active sensing has been shown by \cite{salaris_online_2019, napolitano_gramian-based_2021}. In  \cite{preiss_simultaneous_2018}, trajectories are generated to perform optimal online calibration between GPS and inertial measurement unit (IMU), see also \cite{xu_observability-aware_2023}. In \cite{murali_perception-aware_2019}, for visual-inertial navigation systems, the authors have optimized the duration in which landmarks remain within the field of view. In the context of simultaneous localization and mapping (SLAM), those methods pertain to active SLAM, see \cite{willners_adaptive_2023}. 

One way  to attack active sensing is through the use of  Partially Observable Markov Decision Processes (POMDPs) \cite{kaelbling_planning_1998}, see \cite{candido_minimum_2010}, which offer a proper mathematical framework, but whose complexity is often prohibitory \cite{papadimitriou_complexity_1987}. Sampling-based  planners \cite{he_planning_2008, prentice_belief_2009} may be subject to the same issues. A more tractable option, that we presently adopt, is to work with the widespread extended Kalman filter (EKF). It estimates in real time the state $x_n$ from various sensor measurements, and (approximately) conveys the associated extent of uncertainty  through the state error covariance matrix $P_n$. The magnitude of $P_n$ may then be used as an objective to minimize, as advocated in, e.g.,  \cite{patil_scaling_2015}. 
 
To derive trajectories that minimize $P_n$, a first step is to compute the  gradients of $P_n$ with respect to  the control inputs. To this aim, several approaches are possible. One can use a brute force approach (in the vein of \cite{abbeel_discriminative_2005}) or numerical differentiation, that may be ill-conditioned and intractable, owing to the complexity of the EKF's equations. To get around those problems, \cite{patil_scaling_2015} advocates using backpropagation, through automatic differentiation, and argues that deriving analytical expressions would be difficult. 

We fill a   gap in this paper, by providing novel closed-form analytical expressions over a horizon for the derivatives of any smooth function of the  covariance matrices  of an EKF, w.r.t. all   control inputs, leveraging the powerful backpropagation method. Those   equations  partly extend  our recent results for the linear Kalman filter \cite{parellier_speeding-up_2023} to the nonlinear case, using an EKF. Besides being analytical,   these equations lead to further speedups even over automatic differentiation, as we will show in this paper. We will then use those gradients to solve  perception-aware optimal motion planning. 

Note that some active sensing methods   revolve instead around the Observability Gramian (OG), used to elicit observability   as in \cite{salaris_online_2019}, 
 or as a surrogate for  $P_n$, as advocated in  \cite{cognetti_optimal_2018, salaris_online_2019, napolitano_gramian-based_2021, preiss_simultaneous_2018, bohm_cop_2022}.  Although the covariance matrix and the OG  may be related, see \cite{salaris_online_2019}, or see the probabilistic interpretation of the OG in \cite{benhamou_optimal_2023}, we  prefer to  focus on the covariance matrix, as advocated in  \cite{salaris_online_2019, napolitano_gramian-based_2021, preiss_simultaneous_2018, bohm_cop_2022}, that more realistically captures the noise characteristics, see \cite{rafieisakhaei_use_2017}.


The main contributions of this paper are as follows:
\begin{itemize}
    \item Deriving novel analytical backpropagation equations for the gradient of the covariance of an EKF with respect to all inputs 
 of the filter, including control variables, thus partly extending \cite{parellier_speeding-up_2023} to the  relevant nonlinear context;
 \item Hence providing a computationally efficient and general method whose cost to compute the gradients w.r.t. \emph{all} control inputs at once, over an $N$-step horizon, is similar to that of running one EKF over this horizon, i.e., of complexity $O(Nd^3)$ with $d$ the dimension of the state.
    \item Applying the technique to derive a computationally efficient  perception-aware method;
    \item   Validate and compare the technique through simulations, and real experiments on a  car-size ground vehicle. 
\end{itemize}

 {Section \ref{sec:Primers} introduces the considered problems. Section \ref{sec:backward step} establishes backpropagation equations for computing this gradient. In Section \ref{sec:Path planning formulation}, our path planning formulation is introduced. Finally, we demonstrate the benefits of our approach in simulations (Section \ref{sec:Simulation}) and real experiments (Section \ref{sec:Real-Robot-Experimental}).}

\section{Considered problems}
\label{sec:Primers}

For partially observed linear dynamical systems affected by white Gaussian noise, the Kalman filter (KF) computes the statistics of the state given past observations, namely $p(x_n|y_0,\dots,y_n)$, in real time. The Kalman filter (KF) relies on parameters such as the process noise with covariance $Q_n$ and the observation covariance $R_n$. There exists various approaches to compute the derivatives of the KF's outputs w.r.t. those parameters. The early sensitivity equations \cite{gupta_computational_1974}, see also \cite{tsyganova_svd-based_2017}, allow for computing the  derivative of the likelihood $\mathcal L:=\log p(y_0,\dots,y_n)$ w.r.t. the noise parameters. A much faster approach is to use backpropagation, either using numerical auto-differentiation,  as advocated in \cite{patil_scaling_2015}, or closed-form formulas as very recently derived in  \cite{parellier_speeding-up_2023}.  

In this paper, we  target closed-form backpropagation formulas    for computing the gradients of an EKF's final covariance w.r.t.  its inputs, including the control inputs. This will provide  a nontrivial (partial) extension of the results of \cite{parellier_speeding-up_2023} to nonlinear systems. We start with a few primers.

\subsection{The extended Kalman filter (EKF)}
\label{sec:Preliminaries}
Let us consider a nonlinear discrete-time system:
\begin{equation}
    \left \lbrace
    \begin{aligned}
        &x_n = f(x_{n-1}, u_n, w_n)\,, \quad x_0 = x^0\,,\\
        &y_n = h(x_n) + v_n\,.
    \end{aligned}
    \right .
\label{eq:equation system}
\end{equation}
where $x_n \in \R^\dimx$ is the system's state, $u_n \in \R^\dimu$ is the control input, $w_n \in \R^\dimw$ represents the process noise which follows a Gaussian distribution with zero mean and covariance matrix $Q_n$. The measured output is denoted by $y_n \in \R^\dimy$, corrupted by a Gaussian measurement noise $v_n$ with zero mean and covariance matrix $R_n$.  Owing to unknown noises corrupting the equations, and to the state being only partially observed through $y_n $, one needs to resort to a state estimator. 

The Extended Kalman Filter (EKF) provides a joint estimation of the state, denoted as $\hat{x}_n$, and its covariance matrix, denoted as $P_{n}$, consisting  of two steps.  

At the propagation step, the estimated state is evolved through the noise-free model, that is,  \begin{align}
    \hat x_{n|n-1} = f(\hat x_{n-1|n-1}, u_n, 0),\label{eq:estimate_propagation}
\end{align} and the covariance of the state error   is evolved as
 \begin{align} 
    P_{n|n-1}  &= F_n P_{n-1|n-1} F_n^T + G_n Q_n G_n^T\,.     
    \label{eq:riccati_propagation}
\end{align}
At the update step, the estimated state and the covariance matrix are updated in the light of observation $y_n$ as
\begin{align}
S_n   & = H_n P_{n|n-1}H_n^T + R_n\,,   \label{eq:ups} \\ 
K_n   & = P_{n|n-1}H_n^TS_n^{-1}\,,   \label{eq:upk} \\
\hat x_{n|n}&=\hat x_{n|n-1}+K_n(y_n-h(\hat x_{n|n-1}))\,,\label{eq:state_upp}\\
P_{n|n}  & = (I - K_n H_n)P_{n|n-1}.   \label{eq:riccati_upp}
\end{align}
The Riccati update step \eqref{eq:riccati_upp} proves equivalent to the following update in so-called information form:
\begin{equation}
    P_{n|n}^{-1} = P_{n|n-1}^{-1} + H_n^T R_n^{-1} H_n. \label{eq:information_update}
\end{equation}
In these equations, matrices $F_n$, $G_n$ and $H_n$ are all jacobians that depend on state estimation $\hat{x}_n$ and control inputs $u_n$:
\begin{equation}
    \begin{aligned}
        F_n &= \frac{\partial f}{\partial x}(\hat{x}_{n-1|n-1}, u_n, 0)\,,\quad
        G_n  = \frac{\partial f}{\partial w}(\hat{x}_{n-1|n-1}, u_n, 0)\,, \\
        H_n &= \frac{\partial h}{\partial x}(\hat{x}_{n|n-1}). 
    \end{aligned}    \label{eq:jacobian_prop_F}
\end{equation} 
Through the  Jacobians, the control inputs  affect the covariance matrices, which is in   contrast with the linear case where all trajectories are equivalent. It thus makes sense to compute the sensitivity of covariance matrices w.r.t. control inputs.
 
\subsection{Backpropagation based gradient computation}
\label{sec:backpropagation strategy}
The final covariance computed in the EKF $P_{N|N}$ over a fixed window, say $n=0,\dots,N$, is the result of an iterative algorithm \eqref{eq:estimate_propagation}-\eqref{eq:jacobian_prop_F}, and can thus be viewed as a composition of many functions from $n=0$ up to $N$, as layers in a neural network. As a result, it lends itself to   backpropagation, a way of computing the chain rule backwards. Backpropagation (``backprop") is very efficient when there are numerous inputs and the output is a scalar function (as opposed to the more intuitive method of applying the chain rule forward).   In  \cite{parellier_computationally_2023}, in the context of linear Kalman filtering, it is used to compute the gradient of the negative logarithm of the marginal likelihood (NLL) w.r.t all observations $y_1,\dots,y_N$. In \cite{patil_scaling_2015}, perception-aware trajectory generation is performed using an optimisation-based method where gradients are computed using an automatic differentiation algorithm.



\subsection{Optimal motion planning problem}
\label{sec:loss_design}

Let   $\mathcal{L}$ be a scalar function that reflects the magnitude of its   argument. The considered minimum uncertainty (or active sensing) motion planning problem writes
\begin{equation}
 \left \lbrace
    \begin{aligned}
        &\underset{ u_1,\dots,u_N,x_1,\dots,x_N}{\min} \mathcal{L}(P_{N|N})\\
        \text{subject to} ~&\\
        &\forall n \leq N~x_n = f(x_{n-1}, u_n, 0), \quad x_0 = x_I,\\
        &\forall n \leq N~ P_{n|n-1}  = F_n P_{n-1|n-1} F_n^T + G_n Q_n G_n^T,\\
        &\forall n \leq N~P_{n|n}  = (I - K_n H_n)P_{n|n-1},\\
         &\forall n \leq N~ u_{\min} \leq u_n \leq u_{\max}\\
        \text{with} ~ & F_n  = \frac{\partial f}{\partial x_n}({x}_{n-1}, u_n, 0),\\
        G_n  &= \frac{\partial f}{\partial w_n}({x}_{n-1}, u_n, 0)\,,~ 
        H_n = \frac{\partial h}{\partial x_n}( {x}_{n}).
    \end{aligned}
    \right.
\label{eq:perception-aware path planning problem}
\end{equation}
where $N$ is the time horizon.    The most common choices for $\mathcal L$ to reflect the final uncertainty are the trace $\mathcal L= \trace(P_{N|N})$, used in \cite{rafieisakhaei_use_2017, preiss_simultaneous_2018, patil_scaling_2015},  or the maximum eigenvalue, which has to be regularized using   Schatten's norm, see \cite{salaris_online_2019}. As $\trace(P_{N|N})$  sums the diagonal terms, which are variances expressed in possibly different units and choice of scales, we renormalize the matrix, as suggested in \cite{preiss_simultaneous_2018}, leading to the objective $\mathcal L= \trace \left(P_0^{-1}P_{N|N}\right).$

The difficulty of the (optimal control) problem above stems from the complicated relationship between the controls $u_1,\dots,u_N$ and the final covariance matrix. The rationale in the present paper is to derive an efficient method to compute the  gradients $\dL{u_n}$, and then to perform gradient descent. 

 Although we focus on $P_{N|N}$, our method seamlessly applies to a loss  that also depends on matrices $P_{n|n},~n \leq N.$


\section{Novel sensitivity equations for the EKF}
\label{sec:backward step}
We consider a fixed window, say, $n=0\dots N$, and we seek to differentiate a function $\mathcal L(P_{N|N})$ of the final uncertainty $P_{N|N}$, w.r.t. all the previous EKF inputs, that are, the noise parameters $R_n,Q_n$,  the control inputs $u_n$, for $n=1\dots N$, and the initial values $\hat x_0,P_{0|0}$. They all affect $\mathcal L(P_{N|N})$ in a complicated manner, through the EKF equations \eqref{eq:estimate_propagation} to \eqref{eq:jacobian_prop_F}. For instance a modification of $u_1$, namely $u_1+\delta u_1$ affects initial Jacobians $R_0,G_0,H_0$ in \eqref{eq:jacobian_prop_F}, that in turn affect all subsequent quantities output by the EKF through eq. \eqref{eq:estimate_propagation} to \eqref{eq:information_update}, finally affecting  $\mathcal L(P_{N|N})$ in a non-obvious manner. 

To analytically compute the derivatives, there are two routes. The historical one is to forward propagate a perturbation, say $\delta u_n$, $n\leq N$, through the equations. It is known as the sensitivity equations, and has been done--at least--for the linear Kalman filter in \cite{gupta_computational_1974}, essentially for adaptive filtering. The other route is to compute derivatives backwards, using the backprop method. It is far less straightforward, and has been proposed only very recently, leading to drastic   computation speedups, see \cite{parellier_speeding-up_2023}, in the context of linear systems. In this paper we heavily rely on our prior work \cite{parellier_speeding-up_2023} to go a step further, by accommodating   nonlinear equations, that is, going from the Kalman Filter to the EKF, with the additional difficulty that the Jacobians depend on the estimates. Our end goal being  active sensing, though,  we restrict  ourselves to losses of the form $\mathcal L(P_{N|N})$. We show this additional dependency lends itself to the backprop framework too. We also extend the calculations to  get the derivatives w.r.t. the control inputs (which would not make sense in the linear case as they are null).

\subsection{Matrix derivatives}
\label{sec:Matrix derivatives}

We now explain the methodology, the main steps, and provide the final equations. As deriving closed-form formulas through backward computation  is  non trivial and lengthy, a full-blown mathematical proof can be found in the appendix. 

Our method heavily relies on two ingredients. First, the notion of derivative  of a scalar function w.r.t. a matrix (More detailed can be found in Appendix \ref{appendix_matrix_derivative}),  and the associated formulas based on the chain rule. Then, dependency diagrams which encapsulate how the functions are composed. 
 
Consider $\mathcal{L}(M)$, a scalar function of a matrix defined as $M=f(X, Y)$, where  $X$ and $Y$ are also matrices. $\dL{M}$ denotes the \emph{matrix} defined by $(\dL{M})_{ij}= \frac{\partial \mathcal L}{M_{ij}}$. We   write $\dL{X}$ to denote  $\frac{\partial \mathcal L\circ f}{\partial X}$, otherwise notation would become impractical. 

The chain rule provides rules of calculus for matrix derivatives. We have the following formulas  \cite{petersen__nodate}: 
\begin{equation}
    \begin{aligned}
    \label{eq:matrix_derivative}
        M &= XYX^T &&\Rightarrow  \dL{X} = 2\dL{M} X Y^T\,, \\
        M &= YXY^T &&\Rightarrow  \dL{X} = Y^T\dL{M}Y\,, \\
        M &= X^{-1} &&\Rightarrow  \dL{X} = -M^T\dL{M} M^T\,, 
    \end{aligned}
\end{equation}
In the particular case where $X$ and $M$ are vectors, we have:
\begin{equation}
    \label{eq:vector_derivative}
    M = f(X) \Rightarrow \dL{X} = J^T \dL{M},
\end{equation}with $J$ the Jacobian matrix of $f$ w.r.t. $X$.

When the matrix $X$ depends on a scalar variable $s$, we have the following formula:
\begin{equation}
    \label{eq:chain}
    M = X(s) \Rightarrow \frac{d\mathcal L}{ds} = \trace \left(\frac{d X^T}{d s} \dL{M} \right).
\end{equation}

\subsection{Backprop equations for the involved matrices}\label{sec:backpropagation_eq}
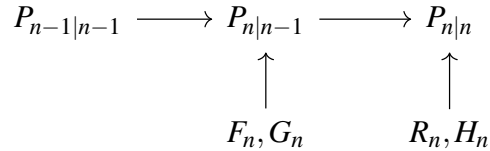
\begin{figure}[!h]
    \centering
    \begin{tikzcd}
        P_{n-1|n-1} \arrow[r] & P_{n|n-1} \arrow[r] & P_{n|n} \\
        & F_n,G_n \arrow[u] & R_n,H_n \arrow[u]
    \end{tikzcd}
    \caption{Dependencies of EKF's variables in Riccati  equations \eqref{eq:riccati_propagation} and \eqref{eq:information_update}. Each variable (node) is a function of its predecessors.}
    \label{fig:graph_dependances_Riccati}
\end{figure}

The graph  in Figure \ref{fig:graph_dependances_Riccati} shows the relationships   involved in the calculation of the state's error covariance, which is our variable of interest.

The backprop method consists in running an EKF until time $N$, to fix the values of all variables, and then compute gradients  backwards to get the derivatives.  
Let us assume we have computed  an expression for $\dL{P_{N|N}}$ (initialization step) at final covariance $P_{N|N}$. It gives how a small variation in  $P_{N|N}$ affects the objective, ignoring all the other variables. Starting with $n=N$, we   go backwards as follows. As $P_{n|n}$ is a function of $P_{n|n-1}$ which is a function of $P_{n-1|n-1}$ in turn, through (\ref{eq:riccati_propagation}) and (\ref{eq:information_update}), we can use formulas (\ref{eq:matrix_derivative}) to assess how a small perturbation in $P_{n|n-1}$ and $P_{n-1|n-1}$ affects the loss in turn (as they affect subsequent quantities, whose variation on $\mathcal L$ has been computed). 
This yields
\begin{align}
        &\dL{P_{n|n-1}}=(I-K_nH_n)^T\frac{\partial \mathcal{L}}{\partial P_{n|n}}(I-K_nH_n)\label{Pnn-1}\,,\\
        &\frac{\partial \mathcal{L}}{\partial P_{n-1|n-1}} = F_n^T \frac{\partial \mathcal{L}}{\partial P_{n|n-1}}F_n\,.
\end{align}

In the same way, we apply formula (\ref{eq:matrix_derivative}) to (\ref{eq:riccati_propagation}) and (\ref{eq:information_update}) to obtain the following relationships: 
\begin{align}
    &\dL{F_n}= 2\dL{P_{n|n-1}}F_nP_{n-1|n-1}\,,\\
    &\dL{G_n} = 2\dL{P_{n|n-1}}G_nQ_n\,,\\
    &\dL{H^T_n}=-2P_{n|n}\dL{P_{n|n}}P_{n|n}H_n^TR_n^{-1}\,, \label{eq:BKprop_H}\\
    &\dL{R_n} = R_n^{-1}H_nP_{n|n}\dL{P_{n|n}}P_{n|n}H_n^TR_n^{-1}\,.\label{eq:BKprop_RR}
\end{align}
Knowing $\dL{P_{n|n}}$ and $\dL{P_{n|n-1}}$, these equations allow in passing to calculate the partial derivative of the loss with respect to the intermediate variables $F_n,~G_n,~H_n$ and $R_n$ at each step.  

\subsection{Backprop equations for the vector variables}
We can now compute the derivatives w.r.t. the state estimates $\hat x$ and the control inputs $u$, which are vectors. However, a remark is necessary. Our end goal is to derive optimal controls $u_1,\dots,u_N$ that minimize loss $\mathcal L(P_{N|N})$. As this is performed ahead of time, the (noisy) observations $y_n$ in \eqref{eq:state_upp} are not available. The most reasonable choice is then to plan using the {\it{a priori}} value of the $y_n$, i.e.,  $y_n=h(\hat x_{n|n-1})$. We may thus alleviate notation writing  $\hat x_{n|n}=\hat x_{n|n-1}:=\hat x_n$ and $\hat x_{n-1|n-1}:=\hat x_{n-1}$.

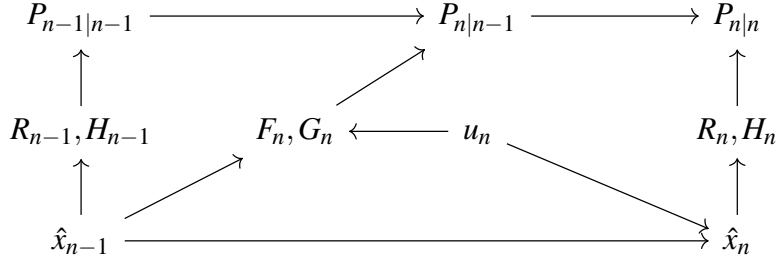
\begin{figure}[!h]
    \centering
    \begin{tikzcd}
        P_{n-1|n-1} \arrow[rr] & & P_{n|n-1} \arrow[rr] & & P_{n|n} \\
         R_{n-1},H_{n-1} \arrow[u] & {F_n,G_n} \arrow[ru] & u_n \arrow[l] \arrow[rrd] & & R_n,H_n \arrow[u] \\
        \hat x_{n-1}  \arrow[u] \arrow[rrrr] \arrow[ru] & & & & \hat x_n \arrow[u]
    \end{tikzcd}
    \caption{Dependency diagram  of all the variables involved in an EKF.} 
    \label{fig:graph_dependances_u}
\end{figure}

The graph in Figure \ref{fig:graph_dependances_Riccati} only focuses on the covariance variables. If we step back, we see the Jacobians depend on the linearization point $\hat{x}_{n-1}$ and the control inputs $u_n$, see \eqref{eq:jacobian_prop_F}. A bigger picture encapsulating all the dependencies in the EKF is represented in   Figure \ref{fig:graph_dependances_u}.

First we compute  $\dL{u_n}$. There is a general rule, derived from the chain rule, which is that $\dL{u_n}$ is the sum of all the derivatives w.r.t. the direct successors of $u_n$ in the graph,   see e.g., \cite{parellier_speeding-up_2023}, provided they have been already computed in a previous step of the backward calculation. 
Additionally using  \eqref{eq:vector_derivative} and \eqref{eq:chain}, and computing  w.r.t. to each scalar component  $u_n^k$  of vector $u_n$, this yields
\begin{equation}
    \dL{u_n^k}= \trace \left(\dL{F_n}^T \frac{\partial F_n}{\partial u_n^k}\right)+\trace \left(\dL{G_n}^T \frac{\partial G_n}{\partial u_n^k}\right)+(J^{u}_ne_k)^T \dL{\hat x_n}\,.  \label{eq:du}
\end{equation}
where $J_n^u := \frac{\partial f}{\partial u}\Bigr|_{\substack{\hat x_{n-1}, u_n,0}}$ with $\hat x_{n-1}, u_n$ computed when running the EKF forward, and $e_k$ the $k$-th vector of the canonical basis (details are given  Appendix \ref{appendix:backprop}).

Similarly, we compute the derivative w.r.t.  the $k$-th scalar component of the system's state $ x_{n-1}^k$,  by adding terms corresponding to each successor in the graph: 
\begin{equation}
    \begin{aligned}
    &\dL{\hat x_{n-1}^k} = \trace\left(\dL{H_{n-1}}^T\frac{\partial H_{n-1}}{\partial \hat x_{n-1}^k}\right) +\trace\left(\dL{R_{n-1}}^T\frac{\partial R_{n-1}}{\partial \hat x_{n-1}^k}\right)\\
    & + (J^x_n e_k)^T \frac{\partial \mathcal{L}}{\partial \hat x_{n}} + \trace\left(\dL{F_n}^T\frac{\partial F_n}{\partial \hat x_{n-1}^k}\right)+  \trace\left(\dL{G_n}^T\frac{\partial G_n}{\partial \hat x_{n-1}^k}\right)\,. 
    \end{aligned}\label{eq20}
\end{equation}
where $J_n^x := \frac{\partial f}{\partial x}\Bigr|_{\substack{\hat x_{n-1}, u_n, 0}}$, which is equal to $F_n$.  

\subsection{Backprop  initialization}
\label{sec:first_step}
To initialize the backward process, one needs to compute $\frac{\partial \mathcal{L} (P_{N|N})}{\partial P_{N|N}}.$  
 In the case of the normalized trace,  we have \cite{petersen__nodate} 
\begin{equation}
    \frac{\partial \mathcal{L} (P_{N|N})}{\partial P_{N|N}}=\frac{\partial  \trace \left(P_0^{-1}P_{N|N}\right)}{\partial P_{N|N}} =  P_0^{-1}. 
\end{equation}
In the case where one targets the maximum eigenvalue of $P_{N|N}$ as a minimization objective, the loss must be chosen consequently. To handle the non differentiability of this cost function, we resort  to its regularized version using Schatten's norm, see  \cite{salaris_online_2019} (details are given in our online preprint \cite{benhamou_backpropagation-based_2024}). 

Note also that at  $n=N$, \eqref{eq20} needs to be adapted, as $\hat x_N$ only has two successors in the graph, thus:\begin{equation}
    \dL{\hat x_N^k}=\trace \left(\dL{H_N}^T\fracd{H_N}{\hat x_{N}^j}\right) + Tr \left(\dL{R_N}^T\fracd{R_N}{\hat x_{N}^k}\right)\label{eq:modif}
\end{equation}




\subsection{Final equations}\label{final:eq:sec}
The gradients may be obtained as follows. We first run the EKF forward, to get all the EKF variables given a sequence of inputs. Then, we may compute the derivative of the loss w.r.t. $P_{N|N}$ at the obtained final covariance matrix. Letting $n=N,$   \eqref{eq:BKprop_H}-\eqref{eq:BKprop_RR}, provide the derivatives w.r.t. $H_N$ and $R_N$, and \eqref{Pnn-1} w.r.t. $P_{N|N-1}$. In turn \eqref{eq:modif} provides the derivative w.r.t. $\hat x_N$, so that \eqref{eq:du} yields the gradient w.r.t. control input $u_N$. Continuing the process backward and using the equations above, we get the derivatives w.r.t. to all control inputs. A summary of the equations is given in Appendix \ref{appendix:summary}.

The process allows for drastic computation speedups, as the  backward equations yield  derivatives w.r.t. \emph{all} control inputs in one pass only. By contrast, forward propagating perturbations would demand running an entire  EKF-like process from scratch for $k=n$ to $ N$ to derive the derivative w.r.t. $u_n$. Akin to dynamic programming, backpropagation allows for reusing  previous computations at each step.

\section{Application to perception-aware planning} 
\label{sec:Path planning formulation}

The computation of the gradients w.r.t. all the EKF's variables  is a contribution in itself that may prove useful beyond active sensing. However, our present goal is to leverage it to address   the  perception-aware optimal  path planning problem \eqref{eq:perception-aware path planning problem}. 
This approach is known as  partial collocation, as in \cite{patil_scaling_2015}, i.e., we explicitly include states as optimization variables and implicitly compute the covariance. A commonly employed method for solving this problem involves a first-order optimization algorithm, the main steps of which are outlined in Algorithm \ref{algo}. From a computational perspective, the gradient computation stands out as the most computationally demanding step, hence the interest for the method developed above.

While \eqref{eq:perception-aware path planning problem} seeks trajectories that minimize the accumulated uncertainty on the robot state over a  given horizon, we note that it is easy to  add constraints or another term in the loss to perform a specific task. For example, adding the constraint $x_N = x_F$ allows for reaching a specific state while being perception-aware. Note it is also easy to make $\mathcal L$ depend on previous $P_{n|n}$, $n<N$, as in \cite{preiss_simultaneous_2018}.

\subsection{Algorithm}
With known gradients,  a \textit{first order} nonlinear optimization algorithm such as Sequential Quadratic Programming (SQP) may be brought to bear. This leads to Algorithm \ref{algo}. 

\begin{algorithm}[h!]
    \caption{Path planning algorithm}\label{algo}
    \begin{algorithmic}
        \Require $x_0,~P_0,~N, \mathcal{L}$
        \State $u \gets (u_0, \dots, u_N)$
        \While{$\mathcal{L}(P_{N|N})$ not converge}
            \State $g \gets $ gradient\textunderscore{}computation$(u)$
            \State $\alpha \gets$ line\textunderscore{}search$(u,g)$ 
            \State $u \gets$  update$(u,g,\alpha)$ \Comment{Using SQP for example}
        \EndWhile
    \State \textbf{Return:} $(u_0, \dots, u_N)$
    \end{algorithmic}
\end{algorithm}
The process involves computing the gradient $g$ of the cost function w.r.t. the control variables using forward and backward passes through the EKF. Additionally, we compute the gradient of the constraints with respect to decision variables. Subsequently, line search is conducted to determine an appropriate step size $\alpha$ for efficient convergence. However, line search requires evaluating the cost function, which corresponds to running a full EKF in our case. The decision variables are then updated using SQP. Further details are given in the experimental sections.

\subsection{Discussion}
In prior work, it proved difficult to use a loss $\mathcal{L}$ depending on the state's covariance $P_{N|N}$. Indeed, the gradient computation of the loss w.r.t each control variable is expensive when using forward difference \cite{preiss_simultaneous_2018, rafieisakhaei_use_2017}, as explained in Section \ref{final:eq:sec}. This has motivated \cite{patil_scaling_2015} to use backpropagation, through automatic differentiation (AD). \cite{patil_scaling_2015} also  argues  deriving analytical formulas would be difficult. 

The interest of our work, that provides analytical formulas, is twofold in this regard.  First, it is often preferable to have closed-form formulas when possible, to rule out many numerical errors and keep better control over the calculation process (possibly opening up for some guarantees about the execution). 
Then, it leads to computation speedups, as it will be shown experimentally in the sequel. 

\section{Simulation results}
\label{sec:Simulation}

We now apply our results to the problem of wheeled robot localization. We consider a 
car-like robot modelled through the unicycle equations, and equipped with a GPS returning position measurements. There are two difficulties associated with this estimation problem. First, the heading is not directly measured. Then, and more importantly, we assume that the position of the GPS antenna in the robot's frame (we call lever arm) is unknown, or inaccurately known, or may slightly vary over time. The resulting problem  pertains to simultaneous self-calibration and navigation, in the vein of \cite{preiss_simultaneous_2018} but in a simpler context. In straight lines, for instance, the lever arm is not observable, so perception-aware trajectories should lead to more accurate robot state estimation. 

In this section we present the model, and we assess and compare our method  through simulations. In the next section,  we apply it to a real world off-road vehicle.

\subsection{Bicycle model}
\label{sec:Simulation model}

The system state consists of the orientation (heading) $\theta_n \in \mathbb{R}$ in 2D, the position of the vehicle $p_n \in \R^2$, and the lever arm $l_n \in \R^2$. The control inputs are the steering angle $\nu_n$ and the forward velocity $\mu_n$.  The kinematic equations based on a roll-without-slip assumption are as follows:
\begin{equation}
    \left \lbrace
    \begin{aligned}
        &\theta_n = \theta_{n-1} + \frac{d t}{L} (\mu_n + w_n^{\mu})\tan(\nu_n + w_n^{\nu}),\\
        &p_n = p_{n-1} + d t \Omega\left(\theta_{n-1} \right)(\mu_n + w_n^{\mu})e_1,\\
        &l_n = l_{n-1}
    \end{aligned}
    \right .
    \label{eq:propagation model simu}
\end{equation}
where $\Omega(\theta)=\begin{pmatrix}
    \cos(\theta)&-\sin(\theta)\\
    \sin(\theta)&\cos(\theta)
\end{pmatrix}$,   $L$ is the distance between the front and the rear wheels, $e_1=(1, 0)^T$ indicates that the velocity is aligned with the robot's heading, and $dt$ is the sampling time. A Gaussian white noise $w_n=(w_n^{\mu}, w_n^{\nu})^T$ with covariance matrix $Q $ corrupts the forward velocity $\mu_n$ and steering angle $\nu_n$ to account for actuators' imperfections, and the mismatch with idealized kinematic model, i.e., slip.  

Letting $l_n\in\mathbb R^2$ be the position of the GPS antenna in the vehicle's frame w.r.t. to its center (the midpoint of the rear axle), the position measured by the GPS is
\begin{equation}
    y_n = p_n + \Omega(\theta_n)l_n + \epsilon_n
    \label{eq:observation model simu}
\end{equation}
where $\epsilon_n$ is a 2D white noise with covariance matrix $R_n$. 

In the simulations, we let $L=4$m,  $d t=1s$. In terms of noise parameters, we let $ Q=\mathrm{diag}(0.1,\pi/180)^2$, and $R_n$ be the identity matrix, i.e., a standard deviation of 1 m for the GPS position measurements.  
To  account for actuator physical limits, we assume $|\nu_n| \leq 30 \pi / 180$ rad and $ 0 \leq \mu_n \leq 5$ m.s$^{-1}$. To account for acceleration limits, we add the constraints $|\Delta \nu_n| \leq 15 \pi/180$ rad.s$^{-1}$ and $|\Delta \mu_n| \leq 1$ m.s$^{-2}$.

\subsection{Simulation results}
\label{sec:Simulation results}

\begin{figure}[h!]
    \centering
    \includegraphics[width=0.98\columnwidth]{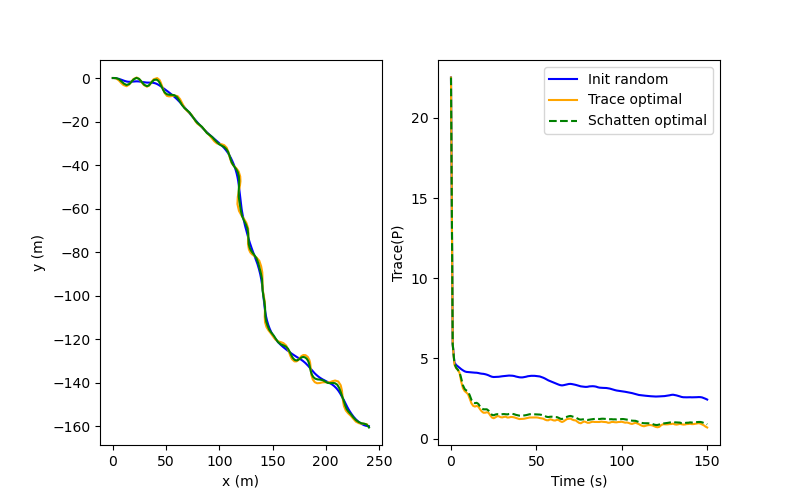}
    \caption{On the left, an example of an initial random guess in blue. The other two trajectories are solutions to the perception-aware problem where the loss is the trace (in orange) and the  Schatten norm (in green). The right plot shows the expected trace of the covariance evolution for each trajectory.}
    \label{fig:traj}
\end{figure}

We start by sampling  admissible control inputs over a  horizon $N=150$s. Then, the corresponding trajectory is obtained by integration. We then use the sequential least-squares programming (SLSQP) algorithm from Scipy \cite{virtanen_scipy_2020} to optimize the sequence of controls to apply. The calculation of the gradient of the loss with respect to the control inputs is performed using the equations detailed in Section \ref{sec:backward step}. An example of the initial random trajectory and the solution of the perception-aware problem can be found in Figure \ref{fig:traj}. The optimal trajectory for the Schatten norm and for the trace are quite similar in this case, and both oscillate around the initial random trajectory. These oscillations are   manoeuvres that increase the observability of the lever arm. They lead to an actual reduction of the expected theoretical error covariance.

\begin{figure}[h!]
    \centering
    \includegraphics[width=0.97\columnwidth]{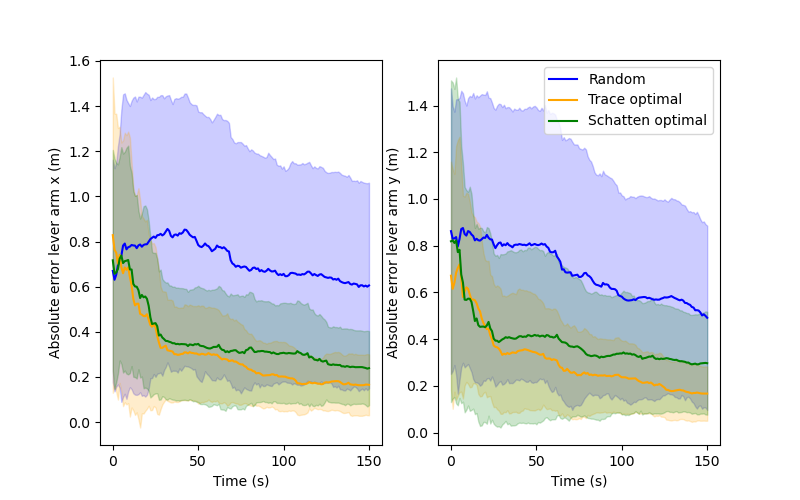}
    \caption{Absolute estimation error of the lever arm during the trajectory. On the left the error for the lever arm in x and on the right in y. One-$\sigma$ envelope illustrates the dispersion of errors over   trials.}
    \label{fig:MAE_lever_arm}
\end{figure}

To demonstrate that the final covariance minimization translates into an actual reduction of the average state estimation error,  we simulate 200 trials of each optimal trajectory by adding process noise and observation noise. During the simulation, the state is estimated with an EKF based on  \eqref{eq:propagation model simu}, \eqref{eq:observation model simu}. The evolution of the absolute estimation error of the lever arm is displayed on  Figure \ref{fig:MAE_lever_arm}. For PA trajectories, the error decreases and converges more rapidly, illustrating the benefit of perception-aware optimal trajectory generation.

\subsection{Computation time}


We compared the computation times for the gradient of the loss w.r.t. control inputs with three different   methods. The first method uses (forward) finite differences to compute the gradient, as in \cite{indelman_towards_2016}. The second method is an  automatic differentiation method, which adopts the backpropagation paradigm, but through a numerical tool, akin to \cite{patil_scaling_2015}. Namely, we used the state-of-the-art PyTorch automatic differentiation (AD) \cite{paszke_pytorch_2019}. Finally, the last method uses our backpropagation analytical formulas of Section \ref{sec:backward step}. The code was executed on computer with an Intel Core i5-1145G7 at 2.60 GHz.

\begin{table}[h!]
    \centering
    \caption{Average and standard deviation of gradient calculation time (over 100 calculations) using different methods.}
    \label{tab:time_computation}
    \begin{tabular}{|l|c|}
    \hline
         Method& Execution time \\
         \hline
         Finite differences&  26.92 $\pm$ 8.45s\\
         PyTorch Autograd& 0.55 $\pm$ 0.19s\\
         Ours & \textbf{0.19} $\pm$ 0.07s\\
         \hline
    \end{tabular}
\end{table}
Table \ref{tab:time_computation} shows that gradient calculation using backpropagation  (Autograd and ours) leads to large speedups. Indeed, when using a finite difference based method, optimization is computationally expensive, as mentioned in \cite{preiss_simultaneous_2018}, where optimizing the trace of the sum of all covariances is reported to take 13 hours for a 3D inertial navigation system. Even when compared to state-of-the-art  PyTorch Autograd, our closed-form formulas are much faster, and more stable in terms of variability of computation time. 

The computation time for the complete resolution of the optimization problem \eqref{eq:perception-aware path planning problem} using our gradient calculation method and Algorithm \ref{algo} is  351s. Although this perfectly suits off-line trajectory generation, it means the robot should stop for a few minutes to plan  in a real-time context.



\subsection{Discussion}

A few remarks are in order. First, we see that, by replacing state-of-the-art autograd differentiation with our formulas, one may  (roughly speaking) double the planning horizon for an identical computation budget. Besides, analytical formulas better suit  onboard implementation. It is interesting to note that they are totally akin to the EKF equations, which must be implemented on the robot anyway. Finally, analytical formulas--when available--may be preferable to numerical methods, as one keeps a better control over what is being implemented, possibly opening up for some guarantees (the behavior of autograd may be harder to anticipate, and leads to higher computation time variability).   Moreover, we anticipate that coding them in C++ may  lead to further speedups.

Note that the overall computation time of the path planning algorithm could be significantly reduced. First, we may obviously optimize over a shorter horizon. Then, the optimizer currently uses a line search algorithm to determine the descent step size. In our case, evaluating the objective function is computationally expensive   because it requires running the EKF. Therefore, finding a method that reduces the number of calls to the objective function  should prove efficient. Finally, another option to reduce computation time is to decrease the number of decision variables, by for instance   parameterizing trajectories using B-splines, as in \cite{salaris_online_2019}.

We may also comment on the obtained trajectories. As the optimization problem is highly nonlinear, non-convex, and constrained,  one should expect an optimization method to fall into a close-by local minimum. In Figure \ref{fig:traj}, the local nature of the optimum proves visible, as the obtained trajectory oscillates around the initial trajectory. Methods to step out of local minima go beyond the scope of this paper. However, it is worth noting that albeit a (close-by) local minimum, the obtained trajectory succeeds in much reducing state uncertainty. It reduces the final average error on the lever arm, which is the most difficult variable to estimate, by a factor 3, see Figure \ref{fig:MAE_lever_arm}. 

In the context of real-time online planning, this suggests a sensible way to use the formulas of the present paper would be to  compute a  real-time trajectory that optimizes a control objective, and then to refine it in real time,  by performing a few gradient descent steps.  This shall (much) increase the information gathered by the sensors.

\section{Real-world experiments}
\label{sec:Real-Robot-Experimental}
 
Real experiments were conducted jointly with the company Safran, a large group that commercializes (among others) navigation systems. With the help from its engineers, we used an    experimental off-road car   owned by the company, which is approximately 4m long and 2.1m wide\footnote{Because of confidentiality requirements, the company has not wished to publish a picture of its experimental vehicle.}. 
\subsection{Experimental setting}
The vehicle is equipped with a standard GPS, odometers, and  a RTK (Real Time Kinematic) GPS, which is not used by the EKF, but  serves as ground-truth for position owing to its high accuracy. The lever arm between the GPS and  the RTK is  denoted by $l_{GPS/RTK}$, and has been calibrated (it is only used for comparison to the ground-truth).

To further test our method, we conducted localization experiments on both ordinary and PA trajectories and compared their performance. We used the model described in Section \ref{sec:Simulation model}, and devised an EKF that fuses odometer data (dynamical model) with GPS position measurements. The vehicle state is represented by a 5-dimensional vector: two dimensions for position, one for the orientation, and two for the lever arm between the GPS and the center of the vehicle frame (midpoint of the rear axle).

Initially, we generated ordinary trajectories at two different speeds (5 $km/h$, 10 $km/h$). Subsequently, we employed our PA path planning Algorithm \ref{algo} initialized with those trajectories, to address optimization problem \eqref{eq:perception-aware path planning problem}, using the final covariance trace as the loss. The constraints used were those relative to the actuators. 
 To demonstrate the ability of optimization-based planning methods to handle further constraints, we also constrained the distance between the initial trajectory and the PA trajectory to be less than 1.5m. Then, we used local tangent plane coordinates computed near the experiment locations to convert 2D planning to world frame coordinates. Then, we followed each trajectory and computed the   localization error committed by the EKF.

To measure localization accuracy, we utilized the RTK-GPS system as a ground-truth, since its uncertainties are of the order of a few centimeters. The position error of the vehicle was computed as follows:
\begin{equation}
    e_n = \left\lVert p_n^{RTK} - \big(\hat{p}_n + \Omega(\hat{\theta}_n)(\hat{l}_n + l_{GPS/RTK})\big) \right\rVert
    \label{eq:error_computation}
\end{equation}

\subsection{Results}
\begin{figure}[!h]
    \centering
    \includegraphics[width=\columnwidth]{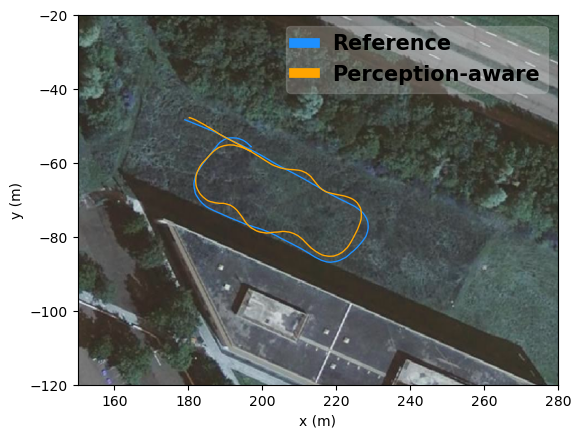}
    \caption{
Off-road trajectories of the RTK-GPS in a local tangent plane coordinates oriented East-North-Up of the scenario at 5 km/h. }
    \label{fig:trajectories_RTK}
\end{figure}

The off-road experiment was performed in a field. It consists of 4 runs: 2 reference runs and 2 perception-aware runs (at 5 $km/h$ and 10 $km/h$). Trajectories returned by  RTK-GPS over the first run are displayed in Figure \ref{fig:trajectories_RTK}. We  note  the oscillations of the PA trajectory around the reference trajectory,  enhancing  state   observability.

\begin{figure}[!h]
    \centering
    \includegraphics[width=0.48\columnwidth]{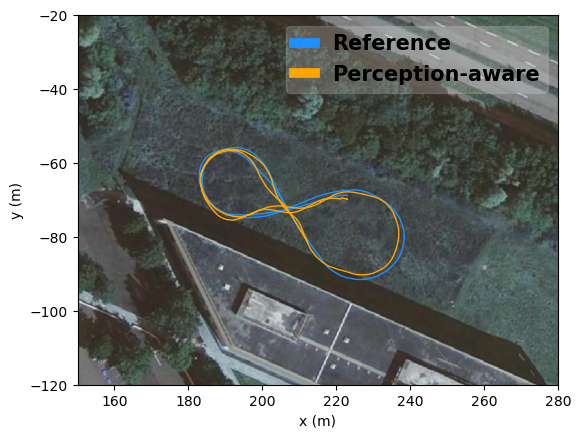}
    \includegraphics[width=0.48\columnwidth]{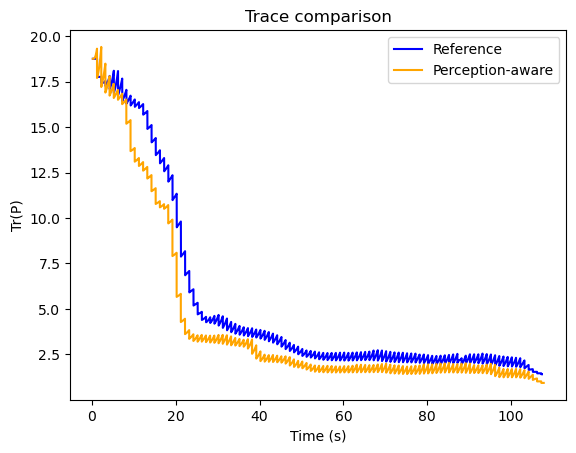}
    \caption{On the left, trajectories of the RTK-GPS in local tangent plane coordinates at 10 km/h. On the right, the evolution of the covariance trace along each trajectory in the scenario at a speed of 10 km/h.  }
    \label{fig:trace_P_vmax10}
\end{figure}

Figure \ref{fig:trace_P_vmax10} shows the trajectories for the second run, along with the evolution of the trace of $P_{n|n}$ output by the EKF (on the first run, the latter is wholly similar and was not included owing to space limitations,   to improve legibility of Fig. \ref{fig:trajectories_RTK}). We see the improvement in terms of trace through the optimized trajectories. The visible wiggling of the covariance corresponds to the correction steps of the EKF. Indeed, the dynamical model is run at 100Hz, corresponding to the odometer frequency, while corrections are made when GPS data are available, at approximately 1Hz.  

\begin{table}[!h]
    \centering
    \caption{Estimation error according to   \eqref{eq:error_computation}   for all trajectories}
    \begin{tabular}{|c|c|c|}
        \hline
        Speed (km/h) & Type & Mean error (m)\\
        \hline
        \multirow{2}{*}{5} & Ref & 3.32\\
        & PA & \textbf{2.88}\\
        \hline
        \multirow{2}{*}{10} & Ref & 4.73\\
        & PA & \textbf{4.34}\\
        \hline
    \end{tabular}
    \label{tab:experiments_res}
\end{table}

Although the trace of the covariance is smaller along the PA trajectory, it doesn't necessarily guarantee that the error with respect to  ground truth is reduced. The uncertainty calculated by the filter assumes all random variables are Gaussian, and the EKF is based on approximations. For each trial, we calculated the average error obtained for each type of trajectory. The results, presented in Table \ref{tab:experiments_res}, point out that PA trajectories enhance localization accuracy. Indeed, in the scenario at 5 km/h, the error decreases by $13.24\%$ between the reference and the PA trajectory, and by $8.25\%$ in the scenario at 10 km/h.

\subsection{Analysis}
When compared with results from simulations, the improvements are   smaller. Several factors may explain this difference. Firstly, in simulations, we calculated an average result using the Monte Carlo method, whereas here we have a single trial. 
Additionally, in simulations, model and noises are perfectly known, whereas in reality, they are not. The roll-without-slip assumption is especially challenged when using an off-road vehicle. This also makes the trajectory tracking quite  imperfect. Finally, due to the limited area, we explicitly constrained the optimized trajectory to stay   around the reference one,  producing a trajectory that optimally ``refines" the reference trajectory, limiting the accuracy increase.

\section{Conclusion}
\label{sec:Conclusion}

Our first contribution has been to introduce  novel backpropagation analytical equations to compute the gradient of any loss based on the covariance of an EKF, w.r.t all its inputs. Beyond the theoretical contribution, they lead to actual numerical speedups. Our second contribution has been to leverage those formulas to address PA path planning, and to test the method in simulation and on real-world experiments over large off-road trajectories over a span of more than 50 meters. In future work, we would like to apply the method to more challenging problems, such as inertial navigation \cite{preiss_simultaneous_2018}, and to improve its scalability, notably by improving the line search. We also would like to combine the method with other  objectives such as reaching a desired goal, collision avoidance,  or energy consumption.

\bibliographystyle{IEEEtran}
\bibliography{Biblio}

\renewcommand{\*}[1]{\mathnormal{#1}}
\renewcommand{\!}[1]{\mathcal{#1}}
\renewcommand{\$}{\partial}

\newcommand{\fracdt}[1]{\frac{\$#1}{\dtheta}}
\newcommand{\fracdl}[1]{\fracd{l_n}{#1}}
\newcommand{\fracdlb}[1]{\fracd{\!{L}}{#1}}
\newcommand{\dv}[2]{{\frac{\partial #1}{\partial #2}}}

\newcommand{\ipriorn}{_{n|n-1}}
\newcommand{\ipostn}{_{n|n}}
\newcommand{\ipostnn}{_{n-1|n-1}}

\newcommand{\xhpn}{\*{\hat{x}}\ipostn}
\newcommand{\xhpnn}{\*{\hat{x}}\ipostnn}
\newcommand{\xhmn}{\*{\hat{x}}\ipriorn}

\newcommand{\Ppn}{\*P\ipostn}
\newcommand{\Ppnn}{\*P\ipostnn}
\newcommand{\Pmn}{\*P\ipriorn}

\newcommand{\dtheta}{\$\theta}
\newcommand{\dK}{\$K}
\newcommand{\Id}{\*I}
\newcommand{\IKH}{\*I - \Kn \Hn}
\newcommand{\Kn}{\*K_n}
\newcommand{\Fn}{\*F_n}
\newcommand{\Bn}{\*B_n}
\newcommand{\Htn}{\*{H}^T_n}

\newcommand{\Sn}{\*S_n}
\newcommand{\Sninv}{\*S_n^{-1}}
\newcommand{\Hn}{\*H_n}
\newcommand{\Rn}{\*R_n}
\newcommand{\Qn}{\*Q_n}
\newcommand{\yn}{\*y_n}
\newcommand{\zn}{\*z_n}

\newcommand{\Lc}{\mathcal{L}}
\newcommand{\lnp}{l\ipostn}
\newcommand{\lnm}{l\ipriorn}
\newcommand{\Linc}{\Lc_{NLL}}

\appendix
\section{Matrix derivatives}
\label{appendix_matrix_derivative}
First, we provide an example of how to derive matrix derivative equations. Then, we will list all the matrix derivative equations we have used in the paper. We consider a scalar function $\Lc : \mathbb{R}^{n\times n} \rightarrow \mathbb{R}$ and we define $\frac{\partial \Lc (X)}{\partial X}$ as the matrix of the derivative of $\Lc$ w.r.t each component of X (i.e. $\left (\frac{\partial \Lc (X)}{\partial X}\right)_{i,j}=\frac{\partial \Lc (X)}{\partial x_{i,j}}$). Using only matrix multiplication, we can express first-order Taylor expansion in the following form (which can alternatively be considered as a definition of the gradient):
\begin{equation}
    \Lc(X+\delta X)= \Lc(X) + Tr\left(\left(\dL{X} \right)^T\delta X\right) + o\left(\delta X \right)
    \label{eq:taylor_expansion}
\end{equation}
Now, let's compute $\frac{\partial \Lc \circ \varphi(X) }{\partial X}$ when $\varphi(X) = XYX^T = M$ expressed as a function of $\dL{M}$.

\begin{equation}
    \begin{aligned}
        \mathcal{L} \circ \varphi(X + \delta X) &= \mathcal{L}((X +\delta X)Y(X +\delta X)^T)\\
        &=\mathcal{L}(XYX^T + XY\delta X^T +\delta X Y X^T + \delta XY\delta X^T )\\
        &= \mathcal{L}(M) + Tr \left( \left(\frac{\partial\mathcal{L}}{\partial M}\right)^T (XY\delta X^T +\delta X Y X^T)\right) + o(\delta X)\\
        &=\mathcal{L}(M) + 2Tr \left( Y X^T\left(\frac{\partial\mathcal{L}}{\partial M}\right)^T \delta X\right)+ o(\delta X)\\
        &=\mathcal{L}(M) + Tr \left( \left(2\frac{\partial\mathcal{L}}{\partial M} X Y^T\right)^T \delta X\right)+ o(\delta X)
    \end{aligned}
\end{equation}
On the other hand, we express the Taylor expansion of $\Lc \circ \varphi$:
\begin{equation}
    \begin{aligned}
        \mathcal{L} \circ \varphi(X + \delta X)&= \mathcal{L} \circ \varphi(X) + Tr \left( \left(\frac{\partial \mathcal{L} \circ \varphi }{\partial X} \right)^T\delta X \right)+ o(\delta X)\\
        &=\mathcal{L}(M) + Tr \left( \left(\frac{\partial \mathcal{L} \circ \varphi }{\partial X} \right)^T\delta X \right)+ o(\delta X)
    \end{aligned}
\end{equation}
By identification we conclude that:
$$ \boxed{\frac{\partial  \mathcal{L} \circ \varphi }{\partial X} = 2\frac{\partial\mathcal{L}}{\partial M} X Y^T} $$\\
which we write more economically as $\frac{\partial  \mathcal{L}   }{\partial X} = 2\frac{\partial\mathcal{L}}{\partial M} X Y^T$, since writing explicitly all the functions composed  in an EKF would be impossible or extremely cluttered. 
Following the same proof structure, we derive:
\begin{equation}
    \begin{aligned}
        \label{eq:matrix_derivative2}
        M &= XYX^T &&\Rightarrow  \dL{X} = 2\dL{M} X Y^T\\
        M &= YXY^T &&\Rightarrow  \dL{X} = Y^T\dL{M}Y \\
        M &= X^{-1} &&\Rightarrow  \dL{X} = -M^T\dL{M} M^T
    \end{aligned}
\end{equation}

Let $A :\mathbb{R}^q \rightarrow \mathbb{M}^{n\times n}$, and we want to compute $\frac{\partial \Lc(A(x))}{\partial x^j}$ ($x^j$ is the $j$-th component of vector $x$) as a function of $\frac{\partial \Lc (A)}{\partial A}$. Employing the Chain rule, we have:

\begin{equation}
    \frac{\partial \mathcal{L} \circ A}{\partial x^j}=Tr \left(\frac{\partial A}{\partial x^j}^T\dL{A}\right) = \sum_{k,l} \left(\frac{\partial A}{\partial x^j}\right)_{k,l} \left(\dL{A}\right)_{k,l}
\end{equation}
Computationally, we favour the first expression.\\

Now, consider a scalar function $\Lc : \mathbb{R}^{n} \rightarrow \mathbb{R}$ and vector-valued function $f: \mathbb{R}^{n} \rightarrow \mathbb{R}^n$  with jacobian $J_x$. We want to compute  $\frac{\partial \Lc \circ f(x)}{\partial x}$ as a function of $\frac{\partial \Lc(m)}{\partial m}$ with $m=f(x)$. Note that  $\frac{\partial \Lc \circ f(x)}{\partial x}$  corresponds then to  the usual (column) vector gradient $\nabla_x(\mathcal L\circ f)$. First, we write the first-order Taylor expansion of $\Lc \circ f$:
\begin{equation}
    \Lc \circ f(x+\delta x) = \Lc \circ f(x) + \langle \nabla \Lc \circ f,\delta x\rangle + o\left( \delta x\right)
\end{equation}

\begin{equation}
    \begin{aligned}
        \mathcal{L}\circ f(x + \delta x) &= \mathcal{L}(f(x) +J_x\delta x + o(\delta x))\\
        &= \mathcal{L}(f(x)) + \langle \nabla \Lc, J_x \delta x\rangle + o(\delta x)\\
         &=\mathcal{L} \circ f(x) + \langle  J_x^T\nabla\Lc,  \delta x\rangle + o(\delta x)
    \end{aligned}
\end{equation}

By identification we conclude that:
\begin{equation}
    \nabla_x (\Lc\circ f) = J_x^T \nabla_m \Lc \Leftrightarrow \frac{\partial \mathcal{L}\circ f(x) }{\partial x} = J_x^T\frac{\partial\mathcal{L}(m)}{\partial m}
\end{equation}

\section{Backpropagation equations in an EKF}
\label{appendix:backprop}
To compute the gradient w.r.t any variable using backpropagation, it is necessary to sum all contributions over the direct successors  of this variable, see, e.g., \cite{parellier_computationally_2023}. The contribution of a direct successor is indicated by an exponent in brackets. Thus, for the direct successors of x denoted by $w_i$, the sum over all contributions is computed as follows:
\begin{equation}
    \dL{x}=\sum_{w_i \text{ direct successor of } x}\left(\dL{x}\right)^{(w_i)}
\end{equation}

For the sake of simplicity, we begin by deriving equations w.r.t. the matrices involved in the Riccati equation. Subsequently, we calculate the contributions w.r.t input control variables and state vector variables.
\subsection{Gradient backpropagation in Riccati equations}

First we consider the propagation step:
\begin{align}  
\Pmn  &= \Fn\Ppnn \Fn^T + G_n\Qn Gn^T\,,     \label{eq:innovp}
\end{align}
followed by an update step (Riccati equation):
\begin{align}
\Sn   & = \Hn\Pmn\Hn^T + \Rn\,,    \\ 
\Kn   & = \Pmn\Hn^T\Sninv\,,   \\
\Ppn  & = (\IKH)\Pmn\ ,\label{recal}\\
\Leftrightarrow \quad \Ppn^{-1}  &  =  \Pmn^{-1}+H_n^TR_n^{-1}H_n\label{recal:info}
\end{align}
We start with a graph showing the relationship between variables to get a better understanding of the dependencies between them.
\begin{figure}[H]
    \centering
    \begin{tikzcd}
        P_{n-1|n-1} \arrow[r] & P_{n|n-1} \arrow[r] & P_{n|n} \\
        & F_n,G_n \arrow[u] & H_n, R_n \arrow[u]
    \end{tikzcd}
    \caption{\centering Dependencies of KF's variables in Riccati's equations}
    \label{fig:graph_dependances_Riccati_sup}
\end{figure}
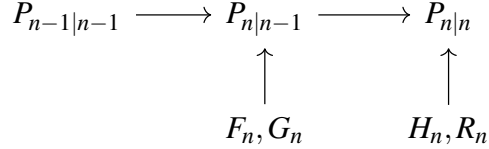
For the posterior covariance, we apply equation (\ref{eq:matrix_derivative2}) to the Ricatti equation. 
\begin{equation}
    \begin{aligned}
        \frac{\partial \Lc}{\partial P_{n-1|n-1}} &= \left( \frac{\partial \Lc}{\partial P_{n-1|n-1}}\right)^{(P_{n|n-1})}\\
        &= \Fn^T \fracdlb{\Pmn} \Fn  
    \end{aligned}
\end{equation}
For the posterior covariance, we depart from the information form of the Riccati equation \eqref{recal:info}
and we find:
\begin{equation}
    \begin{aligned}
        \fracdlb{\Pmn} &=- \Pmn^{-1} \left(\fracdlb{\Pmn^{-1}}\right)^{(\Ppn)} \Pmn^{-1} \\
        &=\Pmn^{-1} \Ppn  \fracdlb{\Ppn}  \Ppn \Pmn^{-1}\\
        &=(\IKH)^T  \fracdlb{\Ppn}   (\IKH), 
    \end{aligned} 
    \label{eq:dp}
\end{equation}where in the last line we have used \eqref{recal}. 
Then we have:
\begin{equation}
    \begin{aligned}
        \fracdlb{\Fn}&=\left(\dL{\Fn}\right)^{(\Pmn)} =2\fracdlb{\Pmn}\Fn\Ppnn
    \end{aligned}
\end{equation}
\begin{equation}
    \begin{aligned}
        \fracdlb{G_n}&= \left(\dL{G_n}\right)^{(\Pmn)}=2\fracdlb{\Pmn}G_nQ_n
    \end{aligned}
\end{equation}

\begin{equation}
    \begin{aligned}
        \left(\dL{H_n}\right)^T &= \left(\dL{H_n}\right)^{(\Ppn)}=2\dL{P_{n|n}^{-1}}H_n^T R_n^{-1}\\
        &= -2P_{n|n}\dL{P_{n|n}}P_{n|n}H_n^T R_n^{-1}
    \end{aligned}
\end{equation}

\begin{equation}
    \begin{aligned}
        \dL{R_n} &= \left(\dL{R_n}\right)^{(\Ppn)} = -R_n^{-1}\dL{R_n^{-1}}R_n^{-1}\\
        &= -R_n^{-1}H_n\dL{P_{n|n}^{-1}}H_n^TR_n^{-1} = R_n^{-1}H_nP_{n|n}\dL{P_{n|n}}P_{n|n}H_n^TR_n^{-1}
    \end{aligned}
\end{equation}

\subsection{Gradient backpropagation with respect to  the  controls}
Assume we have a nonlinear dynamical model:
\begin{align}
    x_n &=f(  x_{n-1} ,\*u_{n}, w_{n}) ,\\
    y_n &=h(x_n) + \nu_n.  
\end{align}
As before we can represent dependencies between variables using a graph:
\begin{figure}[H]
    \centering
    \begin{tikzcd}
        u_{n} \arrow[r] \arrow[rrd] & F_n,G_n & R_n,H_n \\
        x_{n-1} \arrow[ru] \arrow[rr] & & x_n \arrow[u]
    \end{tikzcd}
    \caption{\centering Graph of dependencies of EKF's variables involved propagation and linearization step}
    \label{fig:graph_dependances}
\end{figure}
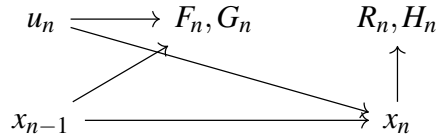

Returning to the Riccati equation, we are interested in using $\mathcal{L}(P_N)$ as a cost function and we would like to assess the sensitivity of $\mathcal L$ w.r.t. the controls $u_n$'s. The influence of the controls on $\mathcal{L}$ is mediated through the parameters $F_n$, $G_n$ $H_n$ and $R_n$. Particularly, with non-linear dynamic, the elements of matrices $F_n$ and $G_n$ may exhibit dependence on both $x_{n-1}$ and $u_n$. Furthermore, $H_n$ and $R_n$ depend on $x_n$. Additionally, it is assumed that the remaining parameters in the Riccati equation remain independent of $x_n$ and $u_n$.\\

Note that, it proves easier  to ignore the relation between function $f$ and matrices $\Fn$. We just see the $\Fn$'s as arbitrary functions of the variables $x_{n-1},u_{n-1}$ for now, allowing for a generic use of the notation of Jacobians that may avoid confusion.\\

By backpropagating the equations of the above subsection over the horizon, we have computed the sensitivity of $\mathcal L$ w.r.t. the various matrices at play. We  now view $\Lc$ as a function of these matrices: 
$$
\mathcal L(F_1,\dots,F_N,G_1,\dots,G_N, H_1,\dots,H_N, R_1,\dots,R_N)
$$with previously computed (known) gradients $  \fracdlb{\Fn}$, $  \fracdlb{G_n}$, $\fracdlb{R_n}$ and $  \fracdlb{\Hn}$. Let us replace the dynamical model above with a linearized model:
\begin{equation}
    \left\lbrace \begin{aligned}
        &x_n=f(x_{n-1}, u_n, w_n)\\
        &\Tilde{x}_n = f(\Tilde{x}_{n-1}, u_n, 0)
    \end{aligned} \right .
\end{equation}
\begin{equation}
    \begin{aligned}
      \Rightarrow  e_n&=x_n -\Tilde{x}_n = f(x_{n-1}, u_n, w_n) - f(\Tilde{x}_{n-1}, u_n, 0)\\
        &\simeq f(x_{n-1}-\Tilde{n-1}+\Tilde{x}_{n-1}, u_n, 0) + G_n w_n - f(\Tilde{x}_{n-1}, u_n, 0)\\
        &\simeq J^x_n (x_{n-1}-\Tilde{x}_{n-1}) + G_n w_n\\
        &=J^x_n e_{n-1}+ G_n w_n
    \end{aligned}
\end{equation}
which is justified as all calculations are to the first order around a nominal trajectory. We denote by $J_u$ the jacobian w.r.t. $u$ and $J_x$ the jacobian w.r.t. $x$ (albeit matrix $F$, but it is less confusing to derive the gradients in the more general case where $f$ is generic with generic jacobian $J_x$).\\

Using this linearization, we calculate the sensitivity of $\mathcal{L}$ w.r.t the component $k$ of $x_{n-1}$:
\begin{equation}
    \begin{aligned}
        \dL{x_{n-1}^k} &= \left(\dL{x_{n-1}^k}\right)^{(F_n)} +\left(\dL{x_{n-1}^k}\right)^{(G_n)} + \left(\dL{x_{n-1}^k}\right)^{(x_n)} + \left(\dL{x_{n-1}^k}\right)^{(H_{n-1})} + \left(\dL{x_{n-1}^k}\right)^{(R_{n-1})}\\
        &=Tr\left(\dL{F_n}^T\frac{\partial F_n}{\partial x_{n-1}^k}\right)+ Tr\left(\dL{G_n}^T\frac{\partial G_n}{\partial x_{n-1}^k}\right) +(J^x_n e_k)^T \frac{\partial \mathcal{L}}{\partial x_{n}}+Tr\left(\dL{H_{n-1}}^T\frac{\partial H_{n-1}}{\partial x_{n-1}^k}\right)\\
        &\quad \quad +Tr\left(\dL{R_{n-1}}^T\frac{\partial R_{n-1}}{\partial x_{n-1}^k}\right)
    \end{aligned}
\end{equation}
where $e_k$ the $k$-th base vector.\\

In the same way, we calculate the derivative of $\Lc$ w.r.t the component $k$ of $u_{n}$:
\begin{equation}
    \begin{aligned}
        \dL{u_{n}^k} &= \left(\dL{u_{n}^k}\right)^{(F_n)} +\left(\dL{u_{n}^k}\right)^{(G_n)} + \left(\dL{x_{n-1}^k}\right)^{(x_n)}\\
        &=Tr\left(\dL{F_n}^T\frac{\partial F_n}{\partial u_{n}^k}\right)+ Tr\left(\dL{G_n}^T\frac{\partial G_n}{\partial u_{n}^k}\right) +(J^{u}_n e_k)^T \frac{\partial \mathcal{L}}{\partial x_{n}}
    \end{aligned}
\end{equation}

\section{First backpropagation step}
\label{appendix:first_step}
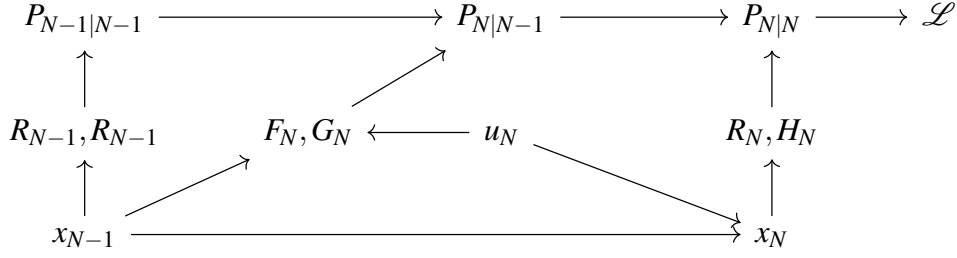
\begin{figure}[H]
    \centering
    \begin{tikzcd}
        P_{N-1|N-1} \arrow[rr] & & P_{N|N-1} \arrow[rr] & & P_{N|N} \arrow[r] & \mathcal{L} \\
        {R_{N-1}, R_{N-1}} \arrow[u] & {F_N,G_N} \arrow[ru] & u_N \arrow[l] \arrow[rrd] & & {R_N,H_N} \arrow[u] & \\
        x_{N-1} \arrow[rrrr] \arrow[u] \arrow[ru] & & & & x_N \arrow[u] & 
    \end{tikzcd}
    \caption{\centering Graph of dependencies of EKF's variables involved in the last step N}
    \label{fig:graph_dependencies_N}
\end{figure}

After the forward pass, the backpropagation begins by computing gradients at the last time step N. In this initial backpropagation step, dependencies between variables exhibit some differences, implying distinct equations. To analyze all dependencies, we refer to the graph in Figure \ref{fig:graph_dependencies_N}.

At the last step,  $P_{N|N}$ has only $ \mathcal{L}$ as a sucessor and $x_{N}$ has $H_N$ and $R_N$ (at the step n, $x_n$ has 5 successors : $H_n$, $R_n$, $F_{n+1}$, $G_{n+1}$ and $x_{n+1}$). The partial derivative $\dL{P_{N |N}}$ depends on the $\mathcal{L}$ chosen. For example if we use the trace then $\dL{P_{N |N}}=I_d$. 

For the normalized trace we have:
\begin{equation}
    \frac{\partial \mathcal{L} (P_{N|N})}{\partial P_{N|N}}=\frac{\partial  \trace \left(P_0^{-1}P_{N|N}\right)}{\partial P_{N|N}} =  P_0^{-T}\frac{\partial  \trace \left(P_{N|N}\right)}{\partial P_{N|N}}=P_0^{-1}. 
\end{equation}

For the Schatten's norm defined by:
\begin{equation}
    ||P_{N|N}||_{\mu} = \left(\sum_i \lambda_i^{\mu}\right)^{1/\mu} = \left(\sum_i (v_i^T \lambda_i  v_i)^{\mu}\right)^{1/\mu}
\end{equation}
where $\lambda_i(P)$ is the i-th eigenvalue of $P_{N|N}$ associated with the eigenvector $v_i$. When the parameter $\mu \gg 1$, then Schatten's norm approximates the highest eigenvalue of the matrix P. Then the gradient is computed by:
\begin{equation}
\begin{aligned}
    \frac{\partial ||P_{N|N}||_{\mu}}{\partial P_{N|N}} &=  \frac{\partial \left(\sum_i \lambda_i^{\mu}\right)^{1/\mu}}{\partial P_{N|N}}= \frac{1}{\mu}\left(\sum_i \lambda_i^{\mu}\right)^{\frac{1-\mu}{\mu}}\frac{\partial \sum_i \lambda_i^{\mu}}{\partial P_{N|N}}\\
    &=\frac{1}{\mu}\left(\sum_i \lambda_i^{\mu}\right)^{\frac{1-\mu}{\mu}}\frac{\partial \sum_i \lambda_i^{\mu}}{\partial P_{N|N}}=\frac{1}{\mu}\left(\sum_i \lambda_i^{\mu}\right)^{\frac{1-\mu}{\mu}}\mu\sum_i\left(\lambda_i^{\mu-1}\frac{\partial \lambda_i}{\partial P_{N|N}}\right)\\
    &=\left(\sum_i \lambda_i^{\mu}\right)^{\frac{1-\mu}{\mu}}\sum_i\left(\lambda_i^{\mu-1}\frac{\partial v_i^TP_{N|N}v_i}{\partial P_{N|N}}\right)= \left(\sum_i \lambda_i^{\mu}\right)^{\frac{1-\mu}{\mu}}\left(\sum_i \lambda^{\mu-1} v_i v_i^T\right)
\end{aligned}
\end{equation}

Using $\dL{P_{N|N}}$ we can compute $\dL{x_N}$ by using the 3 following equations: \\

\begin{equation}
    \left(\dL{H_N}\right)^T = -2P_{N|N}\dL{P_{N|N}}P_{N|N}H_N^TR_N^{-1}
\end{equation}

\begin{equation}
    \dL{R_N} = R_N^{-1}H_NP_{N|N}\dL{P_{N|N}}P_{N|N}H_N^TR_N^{-1}
\end{equation}

\begin{equation}
    \fracdlb{x_N^j}=Tr \left(\fracdlb{H_N}^T\fracd{H_N}{x_{N}^j}\right) + Tr \left(\fracdlb{R_N}^T\fracd{R_N}{x_{N}^j}\right)
\end{equation}

\section{Summary}
\label{appendix:summary}
The gradient computation using backpropagation consists of two parts. The first part named forward pass computes all quantities involved in an Extended Kalman Filter, such as $x_n,~P_{n|n},~P_{n|n-1},~F_n,~G_n$ etc. running the filter equations. At each iteration, we set the innovation $z_n=0$ corresponding to the measurement having its predicted value. This justifies our notation of $x_n$ only, in place of $x_{n|n}$ or $x_{n|n-1}$, or even  $\hat x_{n|n}$ or $\hat x_{n|n-1}$. Additionally, during this forward step, we may compute derivatives of direct functions of the variables at play, such as $\frac{\partial F_n}{\partial u_n^k}$ or $\frac{\partial F_n}{\partial x_n^k}$ etc.

Then, in the backward step, gradients are computing recursively. To initialize the gradients, we first compute $\dL{P_{N|N}}$ and $\fracdlb{x_N^j}=Tr \left(\fracdlb{H_N}^T\fracd{H_N}{x_{N}^j}\right) + Tr \left(\fracdlb{R_N}^T\fracd{R_N}{x_{N}^j}\right)$ as explained in Section \ref{sec:first_step}. Then, we recursively use the equations derived in Section \ref{sec:backpropagation_eq} to backpropagate gradients.
\begin{align}
    &\frac{\partial \mathcal{L}}{\partial P_{n-1|n-1}} = F_n^T \frac{\partial \mathcal{L}}{\partial P_{n|n-1}}F_n\\
    &\dL{P_{n|n-1}}=(I-K_nH_n)^T\frac{\partial \mathcal{L}}{\partial P_{n|n}}(I-K_nH_n)\\
    &\dL{x_{n-1}^k} = Tr\left(\dL{F_n}^T\frac{\partial F_n}{\partial x_{n-1}^k}\right)+ Tr\left(\dL{G_n}^T\frac{\partial G_n}{\partial x_{n-1}^k}\right)+ (J^x_n e_k)^T \frac{\partial \mathcal{L}}{\partial x_{n}}\\
    & \quad \quad +Tr\left(\dL{H_{n-1}}^T\frac{\partial H_{n-1}}{\partial x_{n-1}^k}\right) +Tr\left(\dL{R_{n-1}}^T\frac{\partial R_{n-1}}{\partial x_{n-1}^k}\right)
\end{align}

\begin{align}
    &\dL{F_n}= 2\dL{P_{n|n-1}}F_nP_{n-1|n-1}\\
    &\dL{G_n} = 2\dL{P_{n|n-1}}G_nQ_n\\
    &\left(\dL{H_n}\right)^T=-2P_{n|n}\dL{P_{n|n}}P_{n|n}H_n^TR_n^{-1} \\
    &\dL{R_n} = R_n^{-1}H_nP_{n|n}\dL{P_{n|n}}P_{n|n}H_n^TR_n^{-1}
\end{align}
Finally :    
$$ \boxed{\dL{u_n^k}= Tr \left(\dL{F_n}^T \frac{\partial F_n}{\partial u_n^k}\right)+Tr \left(\dL{G_n}^T \frac{\partial G_n}{\partial u_n^k}\right)+(J^{u}_ne_k)^T \dL{x_n} }$$

\end{document}